\documentclass[10pt,twocolumn,letterpaper]{article}

\usepackage{cvpr}              %

\usepackage{multirow}
\usepackage{amsmath}
\usepackage{bm}

\definecolor{cvprblue}{rgb}{0.21,0.49,0.74}
\usepackage[pagebackref,breaklinks,colorlinks,allcolors=cvprblue]{hyperref}

\title{One4D: Unified 4D Generation and Reconstruction via Decoupled LoRA Control}

\author{Zhenxing Mi, Yuxin Wang, Dan Xu\\
The Hong Kong University of Science and Technology (HKUST)\\
{\tt\small zmiaa@connect.ust.hk, ywangom@connect.ust.hk, danxu@cse.ust.hk}
}

\begin{document}
\twocolumn[{
\maketitle
\begin{center}
\vspace{-0.2cm}
\includegraphics[width=0.99\linewidth]{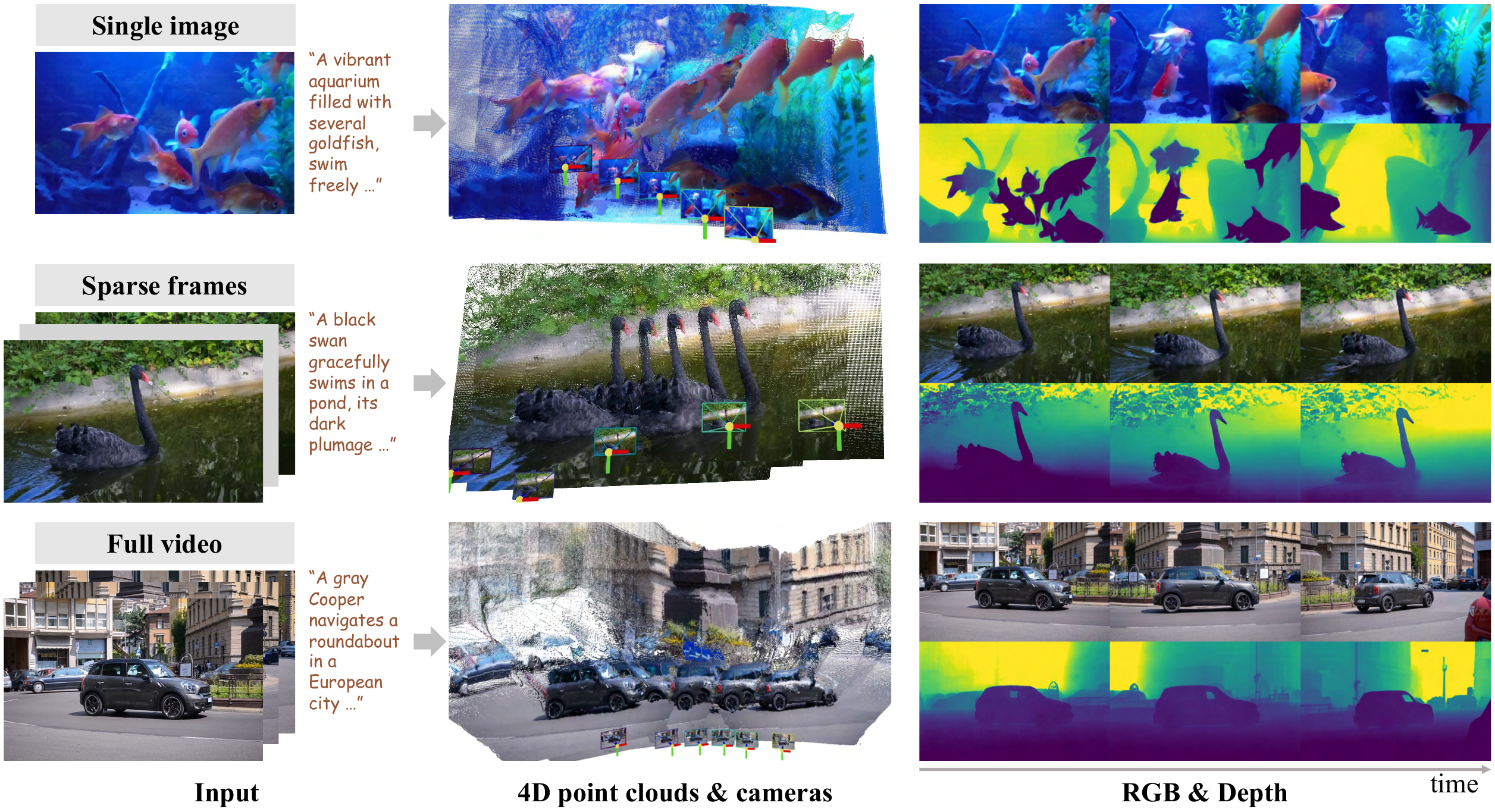}
\end{center}
\vspace{-0.7cm}
\captionsetup{type=figure}
\captionof{figure}{%
One4D supports single-image-to-4D generation, sparse-frame-to-4D generation, and full-video reconstruction in a single model. It outputs synchronized RGB frames and pointmaps, visualized as 4D point clouds with cameras, and RGB-depth sequences.
}\label{fig:teaser}
\vspace{0.4cm}
}]

\begin{abstract}

We present One4D, a unified framework for 4D generation and reconstruction that produces dynamic 4D content as synchronized RGB frames and pointmaps. By consistently handling varying sparsities of conditioning frames through a Unified Masked Conditioning (UMC) mechanism, One4D can seamlessly transition between 4D generation from a single image, 4D reconstruction from a full video, and mixed generation and reconstruction from sparse frames. Our framework adapts a powerful video generation model for joint RGB and pointmap generation, with carefully designed network architectures. The commonly used diffusion finetuning strategies for depthmap or pointmap reconstruction often fail on joint RGB and pointmap generation, quickly degrading the base video model. To address this challenge, we introduce Decoupled LoRA Control (DLC), which employs two modality-specific LoRA adapters to form decoupled computation branches for RGB frames and pointmaps, connected by lightweight, zero-initialized control links that gradually learn mutual pixel-level consistency. Trained on a mixture of synthetic and real 4D datasets under modest computational budgets, One4D produces high-quality RGB frames and accurate pointmaps across both generation and reconstruction tasks. This work represents a step toward general, high-quality geometry-based 4D world modeling using video diffusion models. Project page: \href{https://mizhenxing.github.io/One4D}{\textit{https://mizhenxing.github.io/One4D}}.
\end{abstract}

\vspace{-5pt}
\section{Introduction}
\label{sec:intro}
Simulating the dynamics of the physical world has progressed rapidly with video diffusion and flow-matching models~\cite{StableVideoDiffusion, videoworldsimulators2024, opensora, lin2024open, yang2024cogvideox, genmo2024mochi, wan2025, kong2024hunyuanvideo}. Recent open systems such as Wan~\cite{wan2025}, HunyuanVideo~\cite{kong2024hunyuanvideo}, and Cosmos~\cite{nvidia2025cosmosworldfoundationmodel} demonstrate remarkable visual fidelity and strong understanding of real-world dynamics. However, these foundation models operate purely in RGB space while lacking explicit geometry. Augmenting them with accurate geometry generation is a key step toward downstream world-simulation tasks such as spatial reasoning~\cite{yang2024think}.

Concurrently, scalable 3D/4D foundation models have advanced rapidly.~Dust3R~\cite{dust3r_cvpr24} introduces a pointmap representation encoding both geometry and camera information, enabling efficient feedforward 3D reconstruction. Monst3R~\cite{zhang2024monst3r} and VGGT~\cite{wang2025vggt}, etc., show that pointmaps are effective for static 3D and dynamic 4D reconstruction. Meanwhile, several works extend video or multi-view diffusion models with 6D video representations (RGB+XYZ) for 4D reconstruction~\cite{jiang2025geo4d}, 3D world generation~\cite{zhang2024world}, 3D generation and reconstruction~\cite{szymanowicz2025bolt3d}, and 4D generation~\cite{chen20254dnex}. However, these methods typically specialize in either reconstruction or generation, or are restricted to static 3D scenes.

In this paper, we take a step further and propose One4D, a unified 4D generation and reconstruction framework that significantly enhances the fidelity of joint RGB-geometry modeling through innovative network designs.

We represent each 4D scene as RGB frames and pointmaps due to their flexibility and scalability, where pointmaps are 3-channel 2D (XYZ) videos analogous to RGB videos. Prior work~\cite{jiang2025geo4d, zhang2024world, szymanowicz2025bolt3d, chen20254dnex} shows that modern video VAEs can effectively encode and decode pointmaps. A central challenge in joint RGB-geometry modeling is to fully exploit the strong priors of a pretrained video diffusion model to generate distinct modalities. Existing diffusion-based geometry methods~\cite{ke2023repurposing, jiang2025geo4d, zhang2024world} typically couple RGB and geometry through channel-wise concatenation, which in our setting causes severe cross-modal interference and rapid degradation under low-resource fine-tuning.

To address this, we introduce the Decoupled LoRA Control (DLC), an efficient adaptation design guided by three goals: (i) preserve the base model's strong video priors under low-resource finetuning; (ii) decouple RGB and geometry generation to reduce mutual interference; (iii) enable sufficient cross-modal communication for pixel-level consistency. Concretely, DLC attaches two decoupled and modality-specific LoRA adapters to the video backbone, one for RGB and the other for geometry. The adapters share the frozen base parameters but \emph{not} the base forward computation, forming two \textbf{fully decoupled computation branches}.  The RGB branch maintains pretrained video quality, while the geometry branch adapts to the ``geometry video" distribution. To synchronize the two modalities,  we add zero-initialized control links~\cite{zhang2023adding} between a few corresponding layers across branches. These links gradually learn to transmit information for precise pixel-wise alignment. In practice, DLC prevents mode collapse and yields accurate geometry without sacrificing RGB fidelity.

To unify 4D generation and reconstruction tasks, we further propose Unified Masked Conditioning (UMC). UMC packs different condition types into a single conditioning video that differs in the sparsity of observed frames, filling unobserved frames with zeros. A single-image (with text) corresponds to pure generation. A set of sparse frames corresponds to mixed generation and reconstruction, and a full video corresponds to pure reconstruction. The conditioning video is fed to the diffusion backbone to guide the network to generate missing RGB frames and full pointmaps. With UMC, One4D transitions seamlessly between generation and reconstruction without any architectural changes.

We train One4D on a curated mixture of synthetic and real 4D datasets, combining accurate synthetic geometry and in-the-wild appearance. The resulting model achieves generalizable, high-quality 4D generation and reconstruction in a single framework. Our main contributions are:
\begin{itemize}
    \item We introduce One4D, a unified 4D framework that bridges 4D generation and 4D reconstruction within a single video diffusion model, achieving strong performance across diverse dynamic scenarios.

    \item We propose Decoupled LoRA Control (DLC), which uses modality-specific LoRA branches and zero-initialized control links to preserve video priors while enabling accurate, consistent joint RGB-geometry generation.

    \item We design Unified Masked Conditioning (UMC), a single conditioning interface that supports various 4D generation and reconstruction tasks without model changes.

\end{itemize}

\section{Related Work}
\label{sec:relatedwork}

\noindent \textbf{Video generation models.} Video generation models~\cite{openai2024worldsimulators, opensora, lin2024open, wan2025, kong2024hunyuanvideo, yang2024cogvideox, he2022latent, nvidia2025cosmosworldfoundationmodel} have advanced rapidly with diffusion~\cite{song2020score, ho2020denoising, songdenoising, rombach2022high, peebles2023scalable} and flow-matching models~\cite{lipman2023flowmatchinggenerativemodeling, esser2024scaling}. Modern systems typically combine 3D causal VAE~\cite{kingma2013auto, StableVideoDiffusion, yu2023language, wu2025improved} with DiTs~\cite{peebles2023scalable} to model spatio-temporal dynamics. Recent open-weight models such as Wan~\cite{wan2025}, HunyuanVideo~\cite{kong2024hunyuanvideo}, and Cosmos~\cite{nvidia2025cosmosworldfoundationmodel} demonstrate strong world modeling abilities, suggesting possibilities for geometry generation. Built upon powerful base text-to-video models, image-to-video systems such as Wan~\cite{wan2025} and LongCat-Video~\cite{meituanlongcatteam2025longcatvideotechnicalreport} unify image-to-video, frame interpolation, and video continuation via frame inpainting within a single architecture. One4D follows a similar spirit to unify the 4D generation and reconstruction tasks, with specific designs for joint RGB and geometry generation.

\noindent \textbf{Scalable geometry modeling.} DUSt3R~\cite{dust3r_cvpr24} introduces a paired pointmap representation that encodes both geometry and camera information in a 2D map, enabling scalable 3D reconstruction with transformers. Subsequent works~\cite{zhang2024monst3r, wang2025vggt, tang2025mv, yang2025fast3r, wang2025continuous, jiang2025geo4d} extend this representation to multiview and dynamic reconstruction. Our method leverages these advances by adopting pointmaps as the geometry representation for joint RGB–geometry generation.

\begin{figure}[t]
\centering
    \includegraphics[width=\linewidth]{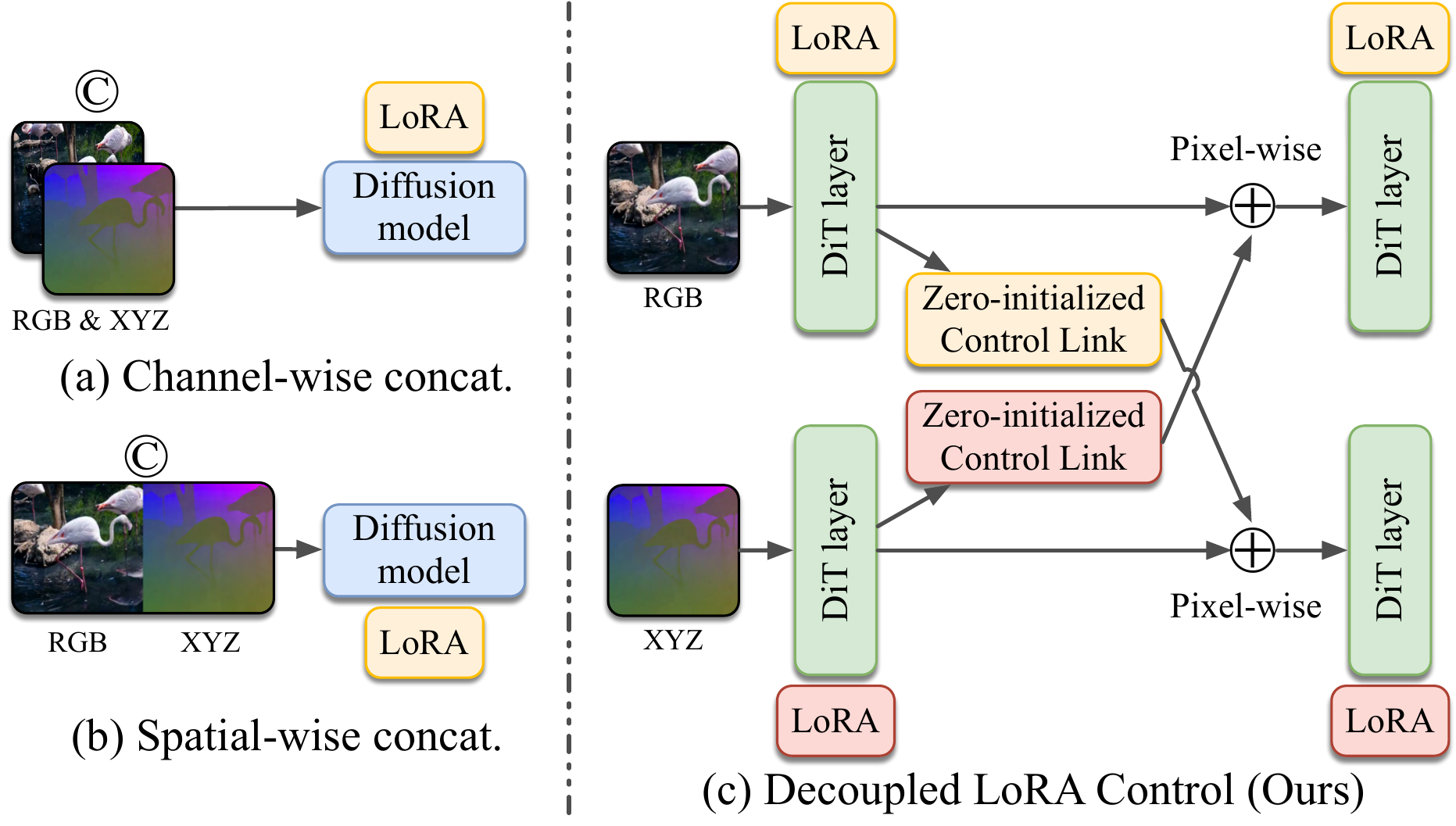}
    \vspace{-0.65cm}
  \caption{Architecture comparison for joint RGB and geometry modeling. (a) Channel-wise and (b) spatial-wise concatenation feed RGB and XYZ into a single diffusion model with a shared LoRA branch. (c) Our Decoupled LoRA Control (DLC) employs two modality-specific LoRA branches with zero-initialized control links, achieving decoupled yet controlled RGB–XYZ joint generation. $\copyright$ denotes concatenation and $\oplus$ denotes pixel-wise addition.
  }
    \vspace{-15pt}
  \label{fig:arch_compare}
\end{figure}

\noindent \textbf{3D and 4D generation.} Leveraging strong 2D diffusion models~\cite{rombach2022high, flux_github} for 3D/4D advances rapidly. Several methods repurpose diffusion models for reconstructing depth and normal maps~\cite{ke2023repurposing, long2024wonder3d}, conditioning noisy geometry with clean RGB latents.
Other works use diffusion models as multiview generators conditioned on cameras, such as CAT3D~\cite{gao2024cat3d}, Bolt3D~\cite{szymanowicz2025bolt3d}, and Stable virtual camera~\cite{zhou2025stable}, and extend this to multi-view videos for 4D generation~\cite{wu2025cat4d, xiesv4d}. Their support for a single or multiple views with camera poses as input is also a multitasking design. Some other works focus on 3D object generation from images or text~\cite{poole2023dreamfusion, zhang2024clay, wu2024direct3d, zhao2025hunyuan3d}.

Several methods model geometry and camera poses jointly via pointmaps, raymaps, or depth maps~\cite{zhang2024world, jiang2025geo4d, chen20254dnex, lu2025matrix3d, szymanowicz2025bolt3d}. WVD~\cite{zhang2024world} generates 6D (RGB+XYZ) static scenes from a single image using channel-wise concatenation of RGB and pointmaps, while 4DNeX~\cite{chen20254dnex} generates dynamic scenes via spatial-wise concatenation. Our method also adopts the RGB+XYZ representation, but differs in two key aspects. (1) We unify dynamic 4D generation and reconstruction within a single model. (2) We introduce Decoupled LoRA Control (DLC), which achieves substantially higher quality and geometry accuracy than channel-wise or spatial-wise concatenation for coupling RGB and geometry.

\begin{figure}[t]
\centering
    \includegraphics[width=\linewidth]{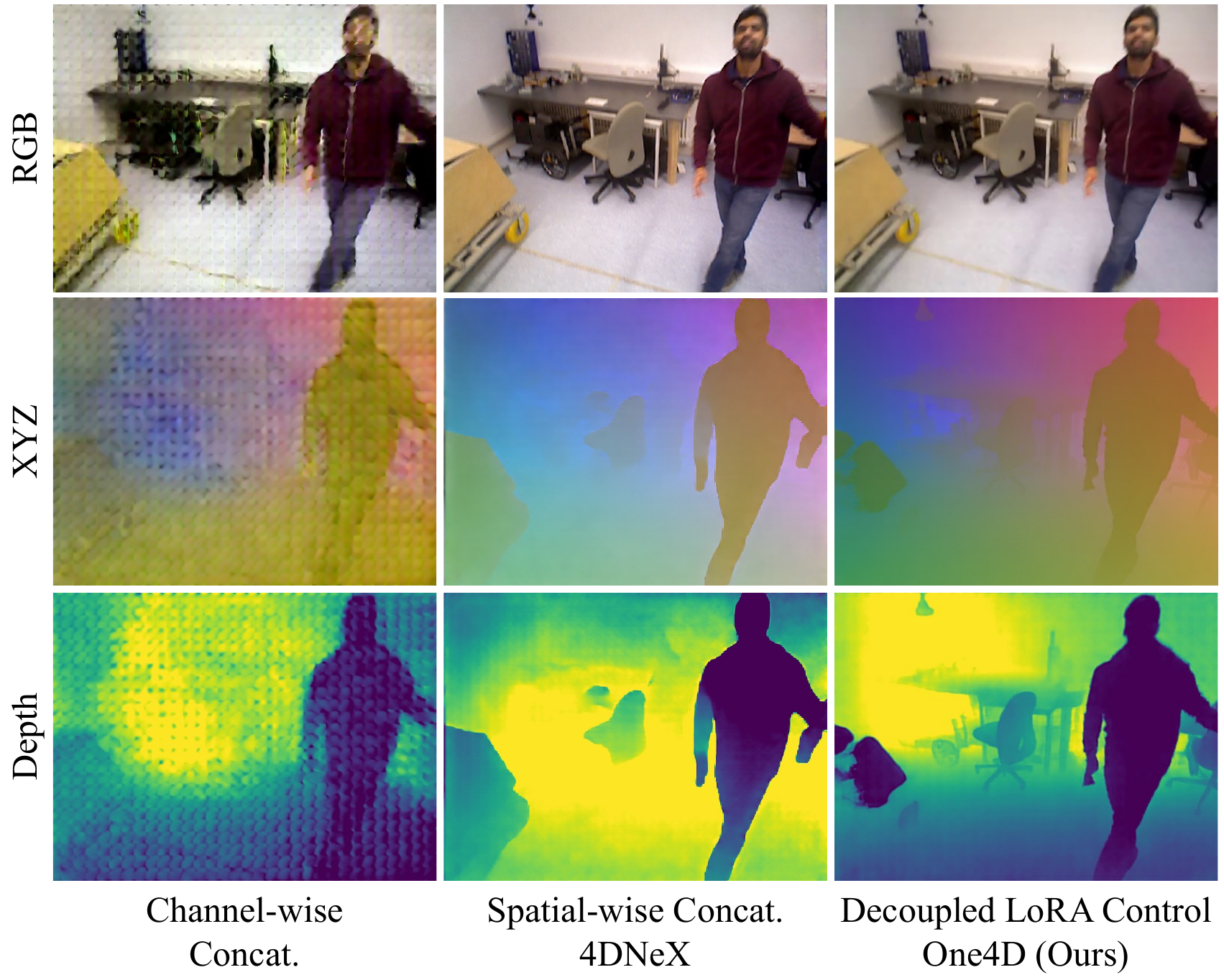}
  \vspace{-0.65cm}
  \caption{Comparison of architectures for joint RGB–geometry generation. Our Decoupled LoRA Control produces cleaner RGB and sharper, more consistent XYZ and depth than channel-wise and spatial-wise concatenation, while channel-wise concatenation severely degrades both appearance and geometry.}

  \label{fig:concat_compare}
\end{figure}

\begin{figure*}[htbp]
\centering
    \includegraphics[width=\linewidth]{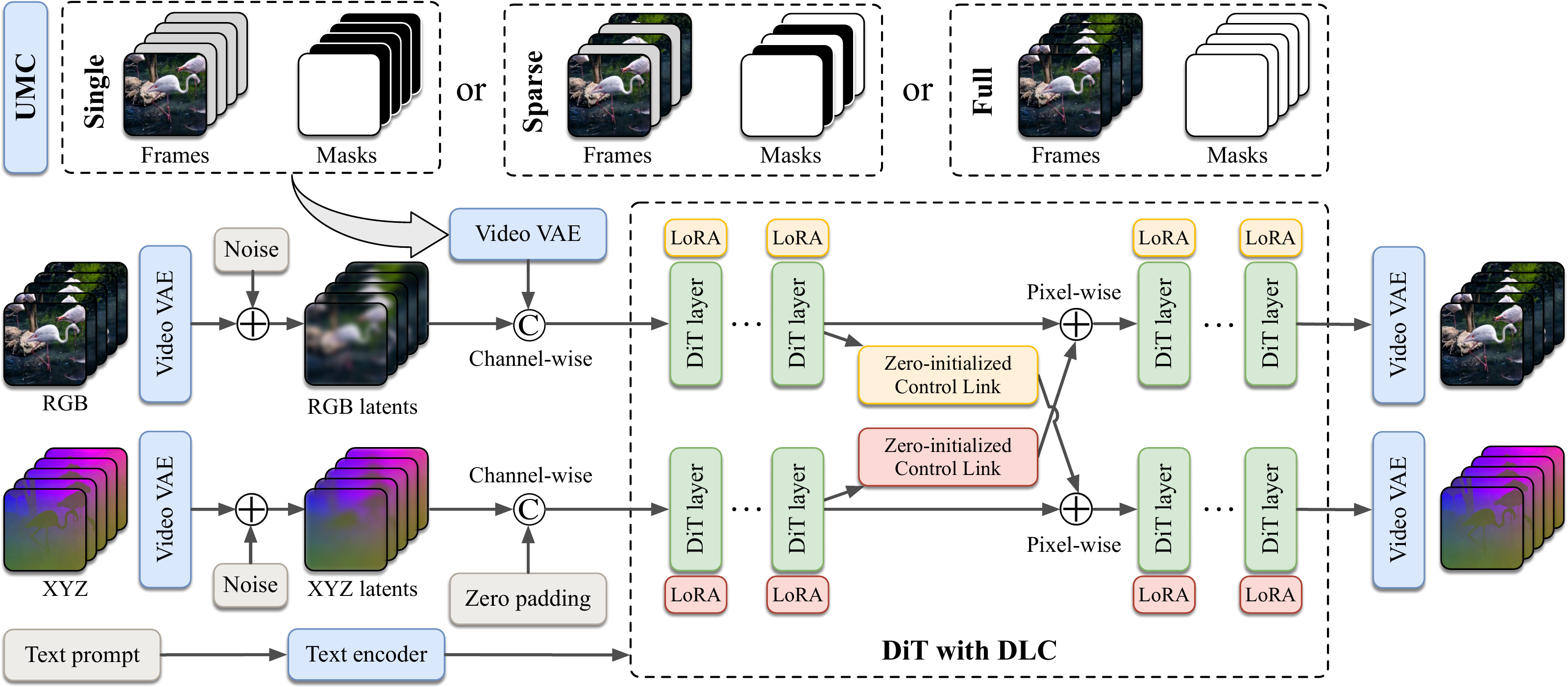}
    \caption{Overview of the One4D framework. Unified Masked Conditioning (UMC) packs single-image, sparse-frame, and full-video inputs into a masked conditioning video. RGB and XYZ videos are encoded into latent spaces via video VAEs, and the conditioning latents are concatenated only with noisy RGB latents. These RGB and XYZ latents are then processed by a DiT backbone with Decoupled LoRA Control (DLC). DLC employs modality-specific LoRA branches to decouple computation, and zero-initialized cross-modal control links to learn pixel-wise consistency. The denoised RGB and XYZ latents are finally decoded into RGB frames and pointmaps.}
  \label{fig:framework}
    \vspace{-10pt}
\end{figure*}

\section{Method}

\subsection{Overview}

One4D extends a flow-matching video generation model~\cite{lipman2023flowmatchinggenerativemodeling, esser2024scaling,wan2025} for joint RGB and geometry generation. As shown in Figure~\ref{fig:framework}, given input images and a text prompt, the model generates synchronized RGB frames $\textbf{X}_{\text{rgb}} \in \mathbb{R}^{3 \times F \times H \times W}$ and pointmaps $\textbf{X}_{\text{xyz}} \in \mathbb{R}^{3 \times F \times H \times W}$, where each pointmap pixel stores the 3D coordinates of the corresponding RGB pixel. $F$ denotes the number of frames and $H\times W$ the spatial resolution. At each training step, RGB and pointmaps are encoded by a video VAE into latents ${\mathbf z}_{\text{rgb}}, {\mathbf z}_{\text{xyz}} \in \mathbb{R}^{c \times f \times h \times w}$. We then sample a timestep $t \in [0,1]$ and Gaussian noise $\epsilon_{\text{rgb}}^t, \epsilon_{\text{xyz}}^t \sim \mathcal{N}(0, I)$, , and construct noisy latents using the Rectified Flow formulation~\cite{lipman2023flowmatchinggenerativemodeling, esser2024scaling}:
\begin{align}
    {\mathbf z}_{\text{rgb}}^t &= t {\mathbf z}_{\text{rgb}} + (1 - t) \epsilon_{\text{rgb}}^t, \label{eq:rf_rgb} \\
    {\mathbf z}_{\text{xyz}}^t &= t {\mathbf z}_{\text{xyz}} + (1 - t) \epsilon_{\text{xyz}}^t. \label{eq:rf_xyz}
\end{align}

The noisy latents and conditioning inputs are fed into DiTs~\cite{peebles2023scalable} to predict velocities supervised by:
\begin{align}  v_{\text{rgb}}^t = {\mathbf z}_{\text{rgb}} - \epsilon_{\text{rgb}} \\
    v_{\text{xyz}}^t = {\mathbf z}_{\text{xyz}} - \epsilon_{\text{xyz}},
\end{align}
and we train with a mean-squared error loss between predicted and ground-truth velocities for both modalities.

Based on this structure, we introduce \emph{Decoupled LoRA Control} (DLC) for stable, modality-specific adaptation with pixel-level consistency between RGB and geometry, and \emph{Unified Masked Conditioning} (UMC) to handle 4D generation and reconstruction in a single model. After generation, camera poses and depthmaps are recovered from the pointmaps via a lightweight global optimization~\cite{dust3r_cvpr24, jiang2025geo4d}.

\subsection{Decoupled LoRA Control}

Previous diffusion-based \textbf{reconstruction} methods, such as Marigold~\cite{ke2023repurposing} and Geo4D~\cite{jiang2025geo4d}, generate depthmaps or pointmaps conditioned on \emph{clean} RGB inputs. They typically concatenate clean RGB latents with noisy geometry latents channel-wise, which works when RGB is fixed. However, in the \textbf{joint generation} setting, both RGB and geometry latents are noisy, and such direct coupling leads to severe quality degradation.  WVD~\cite{zhang2024world} still adopts channel-wise concatenation for joint RGB and XYZ generation, but is trained only on static scenes and requires extreme compute (over 1M steps on 64 A100s). As shown in Figure~\ref{fig:concat_compare}, under moderate compute, we observe that both channel-wise and spatial-wise concatenation~\cite{chen20254dnex} induce premature and excessive interaction between modalities, degrading RGB quality or preventing high-quality geometry learning.

To address these challenges, we propose the Decoupled LoRA Control (DLC), which decouples RGB and XYZ computation to minimize cross-modal interference, while introducing pixel-wise cross-modal communication in a gradually learnable manner.

\noindent \textbf{Decoupled computation.} DLC's first principle is \textbf{decoupling}. We maintain two computation branches for RGB and XYZ tokens, each equipped with its own LoRA adapter, so the model can adapt to each modality without mutual degradation. For a submodule with LoRA in DiTs, the modality-specific computation can be written as:
\begin{align}
    {\mathbf z}_{\text{rgb}}' &= \operatorname{DiTSubmoduleWithRGBLoRA}({\mathbf z}_{\text{rgb}}), \\
    {\mathbf z}_{\text{xyz}}' &= \operatorname{DiTSubmoduleWithXYZLoRA}({\mathbf z}_{\text{xyz}})
\end{align}

Implementing the decoupled computation by simply duplicating all DiT parameters and attaching distinct LoRAs to each copy is infeasible for large-scale base models (14B parameters in our case). Instead, we share the frozen base parameters between modalities and add two separate LoRA adapters on each DiT submodule, while keeping the forward computation disjoint.
\begin{align}
    {\mathbf z}_{\text{rgb}}' &= \operatorname{DiTSubmodule}({\mathbf z}_{\text{rgb}}) + \operatorname{RGBLoRA}({\mathbf z}_{\text{rgb}}), \\
    {\mathbf z}_{\text{xyz}}' &= \operatorname{DiTSubmodule}({\mathbf z}_{\text{xyz}}) + \operatorname{XYZLoRA}({\mathbf z}_{\text{xyz}})
\end{align}
Each DiT submodule is evaluated once per modality, reusing weights but keeping computations decoupled. This drastically reduces memory usage compared to duplicating parameters, making finetuning feasible for large models. The decoupled design of DLC preserves pretrained video priors in RGB branch while allowing XYZ branch to adapt to pointmap generation, achieving high fidelity in both modalities without cross-modal degradation.

\noindent \textbf{Control links.} DLC's second principle is \textbf{control}. For 4D generation, RGB and geometry outputs must be pixel-wise consistent. Channel-wise concatenation enforces strong coupling but harms video quality, while spatial-wise concatenation relies on non-pixel-aligned attention interactions that are relatively weak, as shown in Figure~\ref{fig:concat_compare}. To obtain strong yet controllable cross-modal consistency, we introduce lightweight \emph{control links} to connect the two branches. 

Let the DiT backbone have $N$ layers, and let $\{L_{i_1},\ldots,L_{i_m}\}$, with $1 \le i_1 < \cdots < i_m \le N$ denote a subset of $m$ layers where we insert cross-modal control links. At a linked layer $l$, features of one modality are updated by features from the other modality:
\begin{equation}
\begin{aligned}
\hat{\mathbf z}_{\text{rgb}}^{(l)}
  &= \mathbf z_{\text{rgb}}^{(l)}
   + \operatorname{ZCL}_{\text{rgb}\leftarrow\text{xyz}}\!\bigl(\mathbf z_{\text{xyz}}^{(l)}\bigr),\\
\hat{\mathbf z}_{\text{xyz}}^{(l)}
  &= \mathbf z_{\text{xyz}}^{(l)}
   + \operatorname{ZCL}_{\text{xyz}\leftarrow\text{rgb}}\!\bigl(\mathbf z_{\text{rgb}}^{(l)}\bigr),
\end{aligned}
\end{equation}
where $\operatorname{ZCL}_{\text{rgb}\leftarrow\text{xyz}}$ and $\operatorname{ZCL}_{\text{xyz}\leftarrow\text{rgb}}$ are zero-initialized linear control links. The updated features $\hat{\mathbf z}_{\text{rgb}}^{(l)}$ are $\hat{\mathbf z}_{\text{xyz}}^{(l)}$ then passed to the subsequent DiT layers. Zero initialization~\cite{zhang2023adding} keeps the two branches fully independent at the start of training, preserving the pretrained video priors and the links gradually learn pixel-level alignment between RGB and geometry. In practice, we link only a small subset of layers.

Compared to concatenation-based coupling, DLC provides a controlled, sparsely inserted, and gradually learned pathway for cross-modal communication. As shown in Figure~\ref{fig:concat_compare}, it achieves stronger pixel-wise consistency while maintaining high RGB fidelity and geometric accuracy. It also avoids the token-doubling issue of spatial-wise concatenation within a single attention operation, thereby reducing memory and compute cost.

\subsection{Unified Masked Conditioning}
We introduce Unified Masked Conditioning to express single-image, sparse-frame, and full-video conditioning within a single interface.  Prior video generation models such as Wan~\cite{wan2025} and LongCat-Video~\cite{meituanlongcatteam2025longcatvideotechnicalreport} unify different video generation tasks via frame inpainting. We extend this idea to unified 4D generation and reconstruction, where RGB and geometry are modeled jointly.

\noindent \textbf{Condition construction.} We assemble the available image conditions into a conditioning video $\textbf{X}_{c} \in \mathbb{R}^{C \times F \times H \times W}$, matching the RGB shape, and fill unobserved frames with zeros. $\textbf{X}_{c}$ is encoded by the video VAE into ${\mathbf z}_{\text{c}} \in \mathbb{R}^{c \times f \times h \times w}$, which is concatenated channel-wise with the RGB latents ${\mathbf z}_{\text{rgb}}$. We also build a binary mask $\mathbf{M}_{c} \in \mathbb{R}^{1 \times F \times H \times W}$~\cite{wan2025} indicating observed vs. unobserved frames, reshape it to latent resolution $\mathbf{M}_{c} \in \mathbb{R}^{c_m \times f \times h \times w}$, and concatenate it with RGB latents. The DiT input is:
\begin{equation}
    {\mathbf z}_{\text{input}} = \operatorname{Concat}({\mathbf z}_{\text{rgb}}, {\mathbf z}_{\text{c}}, \textbf{M}_{c})
\end{equation}
Given ${\mathbf z}_{\text{c}}$ and $\mathbf{M}_{c}$, the model will generate missing RGB frames and the full set of pointmaps. This unified construction handles single-image, sparse-frame, and full-video conditioning without changing the architecture.

\noindent \textbf{Controlling geometry.}
Since all XYZ frames are always generated, we do not feed ${\mathbf z}_{\text{c}}$ or $\textbf{M}_{c}$ directly to the geometry branch, avoiding conditioning artifacts in geometry. Instead, conditioning signals reach the geometry branch through DLC control links from the RGB branch, allowing the geometry branch to focus on accurate 3D structure.

With the above designs, UMC makes One4D seamlessly switch between 4D generation and reconstruction tasks within one unified model.

\subsection{Post-Optimization}

After 4D generation, we apply a simple global optimization to recover camera parameters and depth maps from the generated pointmaps, following MonST3R~\cite{zhang2024monst3r} and Geo4D~\cite{jiang2025geo4d}. 
Given the $N$ generated point maps
$\{\hat{\mathbf{X}}^i\}_{i=1}^N$, each
$\hat{\mathbf{X}}^i \in \mathbb{R}^{H \times W \times 3}$, we recover $\{\mathbf{K}^i,\, \mathbf{R}^i,\, \mathbf{o}^i,\, \mathbf{D}^i\}_{i=1}^N,$
where $\mathbf{K}^i$ is the intrinsic matrix, $\mathbf{R}^i$ is the world-to-camera rotation matrix, $\mathbf{o}^i$ is the camera center in the global frame, and $\mathbf{D}^i$ is the depth map of frame $i$.

For each frame $i$ and pixel $(u,v)$, the corresponding 3D point in the global reference frame is parameterized as:
\begin{equation}
\label{eq:x_param_point}
\mathbf{X}_{uv}^i
=
{\mathbf{R}^{i}}^{\!\top}
\big(
D_{uv}^i \,
{\mathbf{K}^i}^{-1}
(u, v, 1)^\top
\big)
+
\mathbf{o}^i ,
\end{equation}
where $D_{uv}^i$ is the depth value at pixel $(u,v)$ in frame $i$.

The predicted point maps $\hat{\mathbf{X}}^i$ is treated as the observations of $\mathbf{X}^i$ and they are aligned by the loss: $\mathcal{L}_{\text{p}}
=
\sum_{i=1}^N
\sum_{u,v}
\left \|
\mathbf{X}^i_{uv} - \hat{\mathbf{X}}^{i}_{uv}
\right \|_1.$
We minimize $\mathcal{L}_\text{p}$ with respect to
$\{\mathbf{K}^i, \mathbf{R}^i, \mathbf{o}^i, \mathbf{D}^i\}_{i=1}^N$, yielding globally consistent camera parameters and depth maps.
We further regularize the camera trajectory by a temporally smooth loss~\cite{zhang2024monst3r, jiang2025geo4d}:
\begin{equation}
\label{eq:align_smooth}
\mathcal{L}_{\text{s}}
(\mathbf{R},
\mathbf{o})
=
\sum_{i=1}^{N-1}
\left(\left \|
{\mathbf{R}^{i}}^{\!\top}
\mathbf{R}^{i+1}-\mathbf{I}
\right \|_{\mathrm{f}}
+\left \|\mathbf{o}^{i+1}
-\mathbf{o}^i\right \|_2\right).
\end{equation}
The final post-optimization objective is a weighted combination of the point-map alignment and trajectory smoothness losses: $\mathcal{L}_{\text {all}}
=
\alpha_1
\mathcal{L}_{\text {p}}
+ \alpha_2 \mathcal{L}_{\text {s}}.$

\begin{figure*}[htbp]
\centering
    \includegraphics[width=\linewidth]{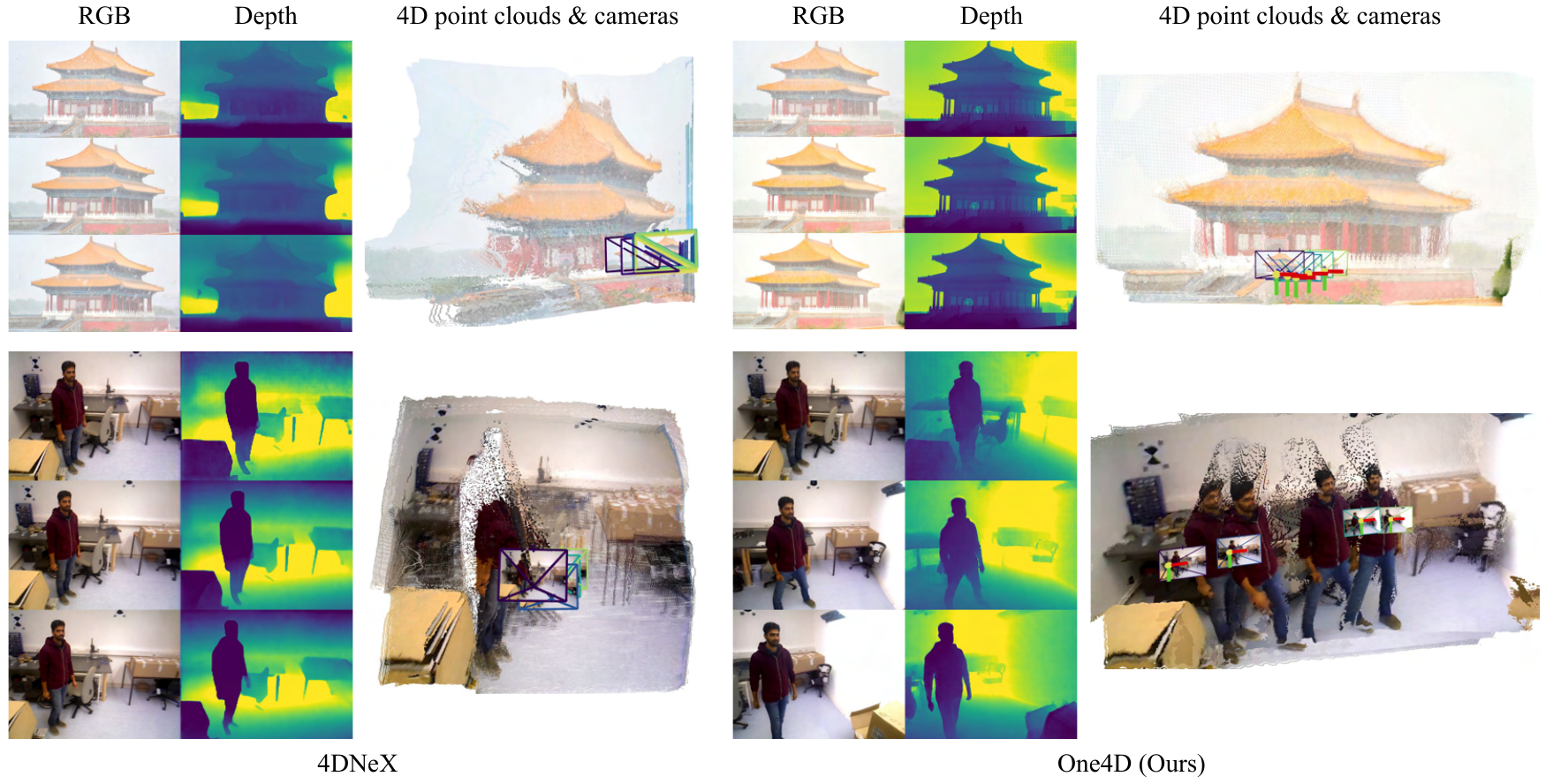}
    \vspace{-20pt}
  \caption{Single-image-to-4D generation comparison between 4DNeX~\cite{chen20254dnex} and our One4D. Compared to 4DNeX, One4D produces more dynamic and realistic videos, sharper and cleaner depth, and more complete, coherent 4D point clouds with cameras.}

  \label{fig:generation_compare}
    \vspace{-15pt}
\end{figure*}

\section{Experiments}

\subsection{Implementation Details}

\noindent \textbf{Datasets.} We construct 4D training datasets from both synthetic and real videos. We first collect dynamic synthetic 4D datasets OmniWorld-Game~\cite{zhou2025omniworld}, BEDLAM~\cite{black2023bedlam}, PointOdyssey~\cite{zheng2023pointodyssey}, TarTanAir~\cite{wangtartanair}, which provide accurate geometry data. To increase dataset scale and diversity, we further annotate real-world videos in SpatialVID~\cite{wang2025spatialvid} with pseudo geometry using Geo4D~\cite{jiang2025geo4d}, covering diverse in-the-wild dynamic scenes. Long videos are clipped into segments of about 81 frames and captioned with Gemini-2.0-Flash~\cite{Google2024Gemini2}. In total, we obtain about 17k synthetic and 17k real-world clips, with roughly 2M frames. Given camera parameters, depth maps are lifted to 3D pointmaps using the first frame as the global coordinate frame, and pointmaps are normalized to $[-1,1]$ before video-VAE encoding.

\noindent \textbf{Training.} One4D is built on Wan2.1-Fun-V1.1-14B-InP~\cite{alibaba2024wan21}, a community finetuned version of Wan2.1-I2V-14B~\cite{wanai2024wan21} for video inpainting. We apply LoRA with rank 64 to all linear layers in the DiT for both RGB and XYZ branches, with 685M parameters. We add DLC control links to five DiT layers, introducing 250.7M parameters. Overall, our model has 935.7M trainable parameters.
Training is done on 8 NVIDIA H800 GPUs with batch size 1 per GPU and gradient accumulation of 4, for 5500 steps at a learning rate of $1 \times 10^{-4}$.
We randomly switch among tasks by varying the number of masked frames. The task sampling ratios are 0.35 for single-image, 0.30 for sparse-frame, and 0.35 for full-video input. The maximum number of training frames is 81, at a resolution of $352 \times 624$.

\noindent \textbf{Inference.} At inference, we use 50 flow-matching steps with a classifier-free guidance scale 6.0. The model jointly generates RGB frames and corresponding pointmaps (XYZ). From the pointmaps, we derive depth maps and estimate camera trajectories via post-optimization. For visualization, pointmaps (XYZ) are interpreted as RGB images. Depth maps are mapped to three-channel images. We use Viser~\cite{yi2025viserimperativewebbased3d} to visualize 4D point clouds together with their camera trajectories, typically subsampling frames with a temporal stride to better show motion.

\begin{figure*}[htbp]
\centering   \includegraphics[width=\linewidth]{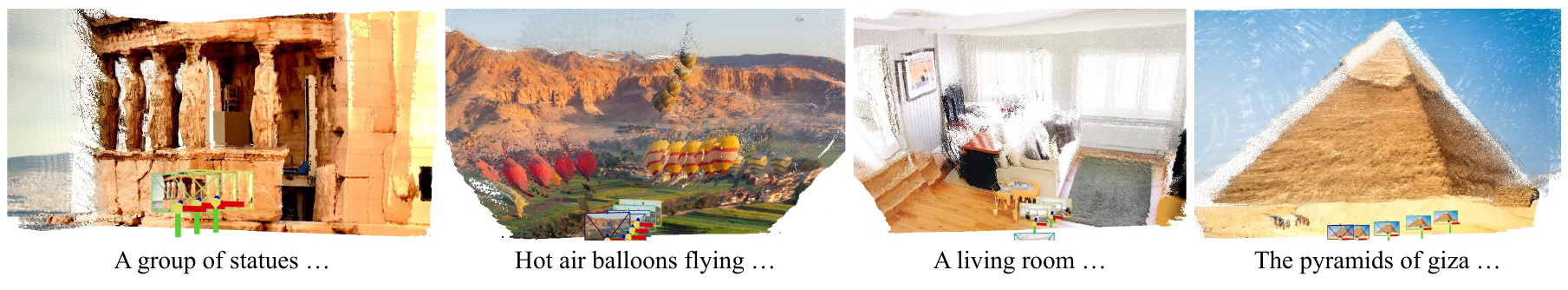}
\vspace{-20pt}  \caption{Additional single-image-to-4D generation results from One4D. It can generate coherent 4D geometry for various types of scenes.}

  \label{fig:more_generation}
  \vspace{-10pt}
\end{figure*}

\begin{table}[t]
  \centering
  \caption{User study comparing 4DNeX~\cite{chen20254dnex} and One4D for 4D generation. Percentages indicate user preference. Our model shows a clear overall advantage, especially on geometry-related criteria.}
  \vspace{-6pt}
  \resizebox{\linewidth}{!}{
    \begin{tabular}{l|ccccc}
    \toprule[1.2pt]
    Method      & Consistency $\uparrow$ & Dynamic $\uparrow$ & Aesthetic $\uparrow$ & Depthmap $\uparrow$ & 4D $\uparrow$ \\
    \midrule
    4DNeX~\cite{chen20254dnex} & 21.0\% & 16.7\% & 17.7\% & 11.7\% & 10.0\% \\
    One4D (Ours)  & \textbf{78.9\%} & \textbf{83.3\%} & \textbf{82.3\%} & \textbf{88.3\%} & \textbf{90.0\%} \\
    \bottomrule[1.2pt]
    \end{tabular}%
    }  \label{tab:userstudygeneration}%
  \vspace{-15pt}
\end{table}%

\subsection{4D Generation}
We compare One4D with the recent 4D generation method 4DNeX~\cite{chen20254dnex}, which relies on spatial-wise concatenation. Both models take a single image and a text prompt as input. We evaluate the generated results using VBench~\cite{huang2024vbench}, a user study, and qualitative comparisons.

As shown in Table~\ref{tab:userstudygeneration}, One4D is preferred over 4DNeX across image-to-video consistency, motion dynamics, aesthetics, and geometry-related criteria such as depth quality and overall 4D coherence. VBench scores in Table~\ref{tab:vbenchforvideoquality} further show that One4D produces stronger, more natural motion while maintaining comparable image-to-video consistency. These results support the effectiveness of our decoupled design, which learns accurate geometry without sacrificing video quality.

\begin{table}[t]
  \centering  \caption{VBench~\cite{huang2024vbench} for video quality. One4D significantly improves motion dynamics and aesthetic quality over 4DNeX, while maintaining comparable image-to-video (I2V) consistency.}
  \vspace{-6pt}
  \resizebox{0.9\linewidth}{!}{
    \begin{tabular}{l|ccc}
    \toprule[1.2pt]
    Method      & Dynamic $\uparrow$ & I2V consistency $\uparrow$ & Aesthetic $\uparrow$ \\
    \midrule
    4DNeX~\cite{chen20254dnex} & 25.6\% & 98.7\% & 61.9\% \\
    One4D (Ours) & \textbf{55.7\%} & 97.8\% & \textbf{63.8\%} \\
    \bottomrule[1.2pt]
    \end{tabular}%
    }  \label{tab:vbenchforvideoquality}%
  \vspace{-15pt}
\end{table}%

Figure~\ref{fig:generation_compare} qualitatively compares RGB frames, 4D point clouds, and depth maps generated by the two methods. One4D generates finer geometric details, more accurate depth, and richer motion, whereas 4DNeX's spatial-wise concatenation struggles to produce fine-grained geometry, as also illustrated in Figure~\ref{fig:concat_compare}. After lifting to 4D, One4D yields coherent scenes with stable backgrounds and naturally moving foreground objects, while 4DNeX often shows limited dynamics. Additional single-image-to-4D results in Figure~\ref{fig:more_generation} demonstrate that One4D produces high-quality geometry and realistic 4D structure across diverse indoor, outdoor, and static, dynamic scenarios.

These high-quality RGB videos and fine-grained, pixel-wise aligned geometry validate the design of Decoupled LoRA Control (DLC), which preserves the strong generative priors of the base model while learning accurate geometry and maintaining pixel-wise consistency.

\begin{table}[t]
  \centering
  \caption{Trained as a unified generation–reconstruction model (G\&R), One4D outperforms reconstruction-only (R) pointmap-based methods such as MonST3R and CUT3R, and remains reasonably close to the reconstruction-only Geo4D-ref, demonstrating effective geometry reconstruction within a unified architecture.}
  \vspace{-6pt}
  \resizebox{\linewidth}{!}{
    \begin{tabular}{lccc|cc}
    \toprule[1.35pt]
      \multirow{2}{*}{Method}  & \multirow{2}{*}{Task} & \multicolumn{2}{c}{Sintel~\cite{butler2012naturalistic}} & \multicolumn{2}{|c}{Bonn~\cite{palazzolo2019refusion}} \\
\cmidrule(lr){3-4} \cmidrule(lr){5-6}
     & & Abs Rel $\downarrow$ & $\delta < 1.25$ $\uparrow$ & Abs Rel $\downarrow$ & $\delta < 1.25$ $\uparrow$  \\
    \midrule
    Marigold~\cite{ke2023repurposing} & R & 0.532 & 51.5  & 0.091 & 93.1 \\
    Depth-Anything~\cite{yang2024depth} & R & 0.367 & 55.4  & 0.106 & 92.1 \\
    NVDS~\cite{wang2023neural}  & R & 0.408 & 48.3  & 0.167 & 76.6 \\
    ChronoDepth~\cite{shao2025learning} & R & 0.687 & 48.6  & 0.100 & 91.1  \\
    DepthCrafter~\cite{hu2025depthcrafter} & R & 0.270 & 69.7  & 0.071 & 97.2 \\
    
    Robust-CVD~\cite{kopf2021robust} & R & 0.703 & 47.8  & -     & -   \\
    CasualSAM~\cite{zhang2022structure} & R & 0.387 & 54.7  & 0.169 & 73.7  \\
    MonST3R~\cite{zhang2024monst3r} & R & 0.335 & 58.5  & 0.063 & 96.4  \\
    CUT3R~\cite{wang2025continuous} &  R & 0.311 & 62.0    &   0.070 & 96.7   \\
    Geo4D-ref~\cite{jiang2025geo4d} & R & 0.205 & 73.5  & 0.059 & 97.2  \\
    \midrule
    One4D (Ours) & G\&R & 0.273 & 70.4 & 0.092 & 93.7  \\
    \bottomrule[1.35pt]
    \end{tabular}%
    }
  \vspace{-5pt}
  \label{tab:reconstructionaccuracy}%
\end{table}%

\subsection{4D Reconstruction}

We evaluate 4D reconstruction in both full-video and sparse-frame settings on several benchmarks to assess our unified generation–reconstruction framework. Sintel~\cite{butler2012naturalistic} provides synthetic video with accurate ground-truth depthmaps, about 50 frames per video. Bonn~\cite{palazzolo2019refusion} contains real dynamic indoor scenes, about 110 frames per video. TUM-dynamics\cite{sturm2012benchmark} contains dynamic scenes sampled to 30 frames for each video. Our evaluation setting closely follows MonST3R\cite{zhang2024monst3r} and Geo4D\cite{jiang2025geo4d}.

We compared depth and camera trajectory derived from our generated pointmaps to reconstruction-only baselines. Marigold~\cite{ke2023repurposing} and Depth-Anything-V2~\cite{yang2024depth} are single-image depth methods. NVDS~\cite{wang2023neural}, ChronoDepth~\cite{shao2025learning} and DepthCrafter~\cite{hu2025depthcrafter} operate on videos. 
In addition, Robust-CVD~\cite{kopf2021robust}, CasualSAM~\cite{zhang2022structure}, MonST3R~\cite{zhang2024monst3r}, CUT3R~\cite{wang2025continuous}, Geo4D~\cite{jiang2025geo4d} jointly reconstruct video depth maps and camera poses.
Although Geo4D~\cite{jiang2025geo4d} is used to annotate our real-world training data, we still report its numbers (Geo4D-Ref) as a strong reconstruction-only reference.

For depth evaluation, we use Sintel~\cite{butler2012naturalistic} and Bonn~\cite{palazzolo2019refusion}, align predicted depths to the ground truth, and report the absolute relative error (Abs Rel) and the percentage of inlier points with $\delta < 1.25$. For camera evaluation, we use Sintel\cite{butler2012naturalistic} and TUM-dynamics\cite{sturm2012benchmark}, and report Absolute Trajectory Error (ATE), Relative Pose Error in translation (RPE-T), and Relative Pose Error in rotation (RPE-R).

\begin{table}[t]
  \centering
  \caption{Camera trajectory accuracy.  The \textit{Task} column distinguishes reconstruction-only (R) methods from our unified generation–reconstruction model (G\&R). One4D achieves camera accuracy comparable to strong reconstruction-only baselines, indicating our unified 4D model performs competitive camera estimation.}
  \vspace{-6pt}
  \resizebox{\linewidth}{!}{
    \begin{tabular}{lcccc|ccc}
    \toprule[1.35pt]
    \multirow{2}{*}{Method}& \multirow{2}{*}{Task} & \multicolumn{3}{c}{Sintel~\cite{butler2012naturalistic}} & \multicolumn{3}{|c}{TUM-dynamics~\cite{sturm2012benchmark}} \\
    \cmidrule(lr){3-5} \cmidrule(lr){6-8}
     & & ATE$\downarrow$  & RPE-T$\downarrow$ & RPE-R$\downarrow$ & ATE$\downarrow$  & RPE-T$\downarrow$ & RPE-R$\downarrow$ \\
    \midrule
    Robust-CVD~\cite{kopf2021robust} & R & 0.360 & 0.154 & 3.443 & 0.153 & 0.026 & 3.528 \\
    CasualSAM~\cite{zhang2022structure} & R & 0.141 & 0.035 & 0.615 & 0.071 & 0.010 & 1.712 \\
    MonST3R~\cite{zhang2024monst3r} & R & 0.108 & 0.042 & 0.732 & 0.063 & 0.009 & 1.217 \\
    CUT3R~\cite{wang2025continuous} & R & 0.208 & 0.062 & 0.610 & 0.046 & 0.014 & 0.446 \\
    Geo4D-ref~\cite{jiang2025geo4d} & R & 0.185 & 0.063 & 0.547 & 0.073 & 0.020 & 0.635 \\
    \midrule
    One4D (Ours) & G\&R & 0.213 & 0.057 & 0.818 & 0.129 & 0.022 & 1.447 \\
    \bottomrule[1.35pt]
    \end{tabular}%
    }
  \label{tab:reconstructioncamera}%
  \vspace{-5pt}
\end{table}%

\begin{table}[t]
  \centering
  \caption{Depth accuracy for sparse-frame-to-4D generation on Sintel and Bonn.~Using only a fraction of frames (Sparsity), One4D remains competitive and degrades gracefully, indicating strong geometry generation from sparse observations.}
  \vspace{-6pt}
  \resizebox{0.86\linewidth}{!}{
    \begin{tabular}{lcc|cc}
    \toprule[1.2pt]
      \multirow{2}{*}{Sparsity}   & \multicolumn{2}{c}{Sintel~\cite{butler2012naturalistic}} & \multicolumn{2}{|c}{Bonn~\cite{palazzolo2019refusion}} \\
\cmidrule(lr){2-3} \cmidrule(lr){4-5}
     & Abs Rel $\downarrow$ & $\delta < 1.25$ $\uparrow$ & Abs Rel $\downarrow$ & $\delta < 1.25$ $\uparrow$  \\
    \midrule
    0.50 & 0.314 & 70.3  & 0.094 & 93.5 \\
    0.25 & 0.443 & 67.7  & 0.094 & 93.3 \\
    0.10 & 0.453 & 64.0  & 0.099 & 92.9 \\
    0.05 & 0.641 & 57.6  & 0.151 & 87.2 \\
    0.04 &    -   &   -    & 0.191 & 82.5 \\
    0.03 &   -    &    -   & 0.277 & 71.1 \\
    \midrule
    \textbf{Full Model}  & \textbf{0.273} & \textbf{70.4}  & \textbf{0.092} & \textbf{93.7} \\
    \bottomrule[1.2pt]
    \end{tabular}%
    }
  \label{tab:sparsegenerationandreconstruction}%
  \vspace{-15pt}
\end{table}%

\begin{figure}[t]
\centering  \includegraphics[width=\linewidth]{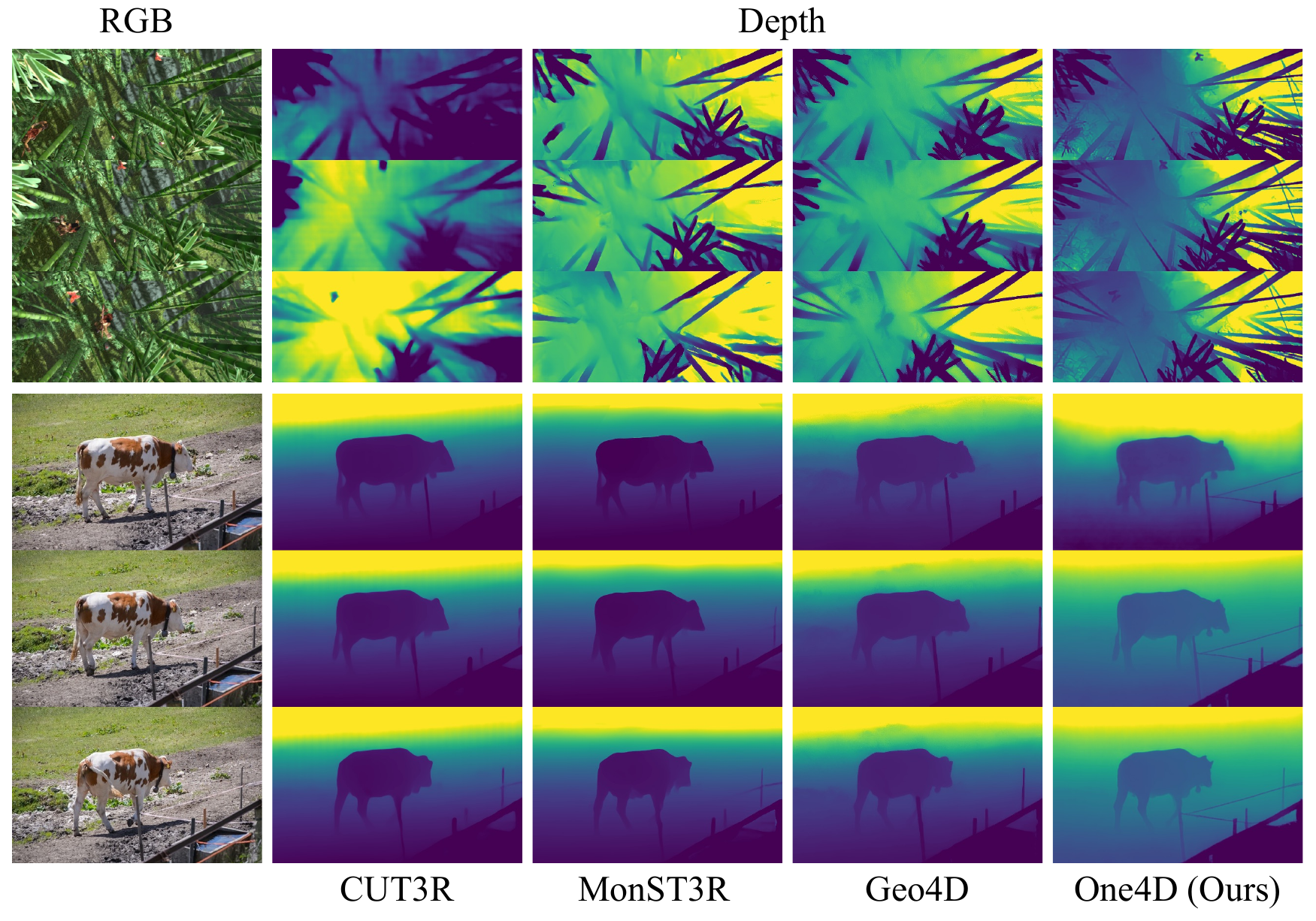}
  \caption{Qualitative full-video 4D reconstruction comparison with CUT3R~\cite{wang2025continuous}, MonST3R~\cite{zhang2024monst3r}, and Geo4D~\cite{jiang2025geo4d}. The proposed One4D recovers sharper object boundaries and more accurate depth, especially on thin structures and challenging geometry.}

  \label{fig:recon_compare}
\end{figure}

\noindent \textbf{Full-video-to-4D.} Table~\ref{tab:reconstructionaccuracy} reports full-video depth accuracy. Among pointmap-based methods, One4D clearly outperforms MonST3R~\cite{zhang2024monst3r} and CUT3R~\cite{wang2025continuous} on Sintel (70.4\% vs 58.5 / 62.0 on $\delta < 1.25$), and remains close to the reconstruction-only Geo4D-ref~\cite{jiang2025geo4d}, even though those models are trained purely for reconstruction and One4D is trained once for both 4D generation and reconstruction. This indicates that our Decoupled LoRA Control and Unified Masked Conditioning allow a single generative model to recover highly accurate geometry. Results for CUT3R are obtained using the official Geo4D evaluation scripts. Results of other baselines are taken from MonST3R~\cite{zhang2024monst3r} and Geo4D~\cite{jiang2025geo4d} papers. 

Camera accuracy in Table~\ref{tab:reconstructioncamera} shows that One4D achieves competitive ATE and RPE scores within the same range of Geo4D-ref and other reconstruction baselines, confirming that our pointmaps are sufficiently accurate to support robust camera estimation. Qualitative results in Figure~\ref{fig:recon_compare} further show that One4D can recover fine geometric structures (e.g., bamboo leaves, ropes) and robustly handles challenging scenes such as dense bamboo forests.

Overall, the strong depth and camera reconstruction results across Sintel, Bonn, and TUM-dynamics highlight that our designs of DLC and UMC enable One4D to effectively reconstruct 4D geometry and camera trajectories, even though we train it as a single unified model for both generation and reconstruction.

\begin{table}[t]
  \centering
  \caption{Ablation on CFG scale and training steps for depth reconstruction. Our model is robust to CFG choice and attains good accuracy with few training steps, improving with longer training.}
  \vspace{-6pt}
  \resizebox{\linewidth}{!}{
    \begin{tabular}{lcc|cc}
    \toprule[1.2pt]
       \multirow{2}{*}{Setting}  & \multicolumn{2}{c}{Sintel~\cite{butler2012naturalistic}} & \multicolumn{2}{|c}{Bonn~\cite{palazzolo2019refusion}} \\
\cmidrule(lr){2-3} \cmidrule(lr){4-5}
     & Abs Rel $\downarrow$ & $\delta < 1.25$ $\uparrow$ & Abs Rel $\downarrow$ & $\delta < 1.25$ $\uparrow$  \\
    \midrule
    CFG=4 & 0.257 & 71.5  & 0.092 & 94.0 \\
    CFG=5 & 0.259 & 70.9  & 0.090 & 94.1 \\
    \midrule
    Step=1000 & 0.331 & 65.4  & 0.114 & 88.9 \\
    Step=3000 & 0.284 & 68.1  & 0.097 & 91.8 \\
    \midrule
    \textbf{One4D (Ours)}  & \textbf{0.273} & \textbf{70.4}  & \textbf{0.092} & \textbf{93.7} \\
    \bottomrule[1.2pt]
    \end{tabular}%
    }
  \label{tab:ablationreconstruct}%
\end{table}%

\noindent\textbf{Sparse-frame-to-4D.}~In the sparse-frame-to-4D setting, we keep the first frame and last frame of each video and uniformly sample additional frames in between, with the total number of observed frames controlled by a sparsity ratio. The model must generate all missing RGB frames and the complete pointmap sequence.

Table~\ref{tab:sparsegenerationandreconstruction} reports depth accuracy on Sintel and Bonn with different spatial ratios. Remarkably, with only $50\%$ or $25\%$ of frames, One4D is almost as accurate as in the full-video setting on both datasets. Even at sparsity $0.10$, degradation is modest. Moreover, under extreme sparsity (e.g., $0.05$ or $0.03$), the model still produces reasonable geometry from only 2 or 3 frames, typically just the first and last frames. This is also partly because these boundary frames already capture sufficient background information. 

Overall, the sparse-frame evaluation shows that our model can reliably generate unobserved RGB frames and corresponding 4D structure from very sparse observations, complementing the single-image-to-4D generation results and highlighting the strength of our unified generation–reconstruction design.

\subsection{Additional Ablation Study}

Beyond our main generation and reconstruction experiments, we further ablate two factors of One4D on Sintel and Bonn, including the classifier-free guidance (CFG) scale and the number of training steps. Results are summarized in Tables~\ref{tab:sparsegenerationandreconstruction} and~\ref{tab:ablationreconstruct}.

We use CFG~$=6$ as the default setting in all main experiments. As shown in Table~\ref{tab:ablationreconstruct}, using CFG~$=4$ or CFG~$=5$ yields very similar depth accuracy, indicating that One4D is robust to the choice of guidance scale. This robustness simplifies deployment in real-world applications.

We also study the effect of training steps by training models with 1K, 3K, and 5.5K optimization steps. Even with only 1k steps, One4D already achieves reasonable accuracy, and at 3k steps, it is close to the full model. Compared to WVD~\cite{zhang2024world}, which requires around 1M steps over two weeks on 64 A100 GPUs with channel-wise concatenation, our decoupled design adapts the base video model with much fewer training steps. This efficiency supports our claim that the proposed architecture preserves and effectively leverages pretrained video priors. As the number of training steps increases, performance generally improves, indicating promising potential for further gains when scaling to larger datasets and more computation.

\section{Conclusion}
\label{sec:conclusion}
We presented One4D, a unified 4D framework that bridges 4D generation and 4D reconstruction within a single video diffusion model. We introduced Decoupled LoRA Control for robust joint RGB and geometry modeling. We proposed Unified Masked Conditioning, a simple conditioning scheme that seamlessly handles pure generation, mixed generation–reconstruction, and pure reconstruction without modifying the architecture. Trained on a curated mixture of synthetic and real 4D data, One4D achieves generalizable, high-quality 4D results and takes a step toward geometry-aware world simulation with video foundation models.

{
    \small
    \bibliographystyle{ieeenat_fullname}
    \bibliography{main}
}

\end{document}